\pdfoutput=1

\documentclass[11pt]{article}

\usepackage[final]{acl}

\usepackage{times}
\usepackage{latexsym}
\usepackage{booktabs}
\usepackage{caption}
\usepackage{tabularx}
\usepackage{adjustbox}

\usepackage[T1]{fontenc}

\usepackage[utf8]{inputenc}

\usepackage{microtype}

\usepackage{inconsolata}

\usepackage{graphicx}

\title{Discrete Diffusion Language Model for Efficient Text Summarization}

\author{
  \makebox[0.5\textwidth][c]{Do Huu Dat\textsuperscript{1,*}\thanks{22dat.dh@vinuni.edu.vn}\thanksref{equalAuth}, Do Duc Anh\textsuperscript{2,*}\thanks{ducanh003@e.ntu.edu.sg}\thanksref{equalAuth}, Anh Tuan Luu\textsuperscript{2}, Wray Buntine\textsuperscript{1}} \\
  \textsuperscript{1}VinUniversity \\
  \textsuperscript{2}Nanyang Technological University, Singapore\\
  \thanks{Equal Contribution}
}

\author{
  \makebox[0.5\textwidth][c]{Do Huu Dat\textsuperscript{1}\thanks{These authors contributed equally to this work.}\thanks{22dat.dh@vinuni.edu.vn}, Do Duc Anh\textsuperscript{2}\footnotemark[1]\thanks{ducanh003@e.ntu.edu.sg}, Anh Tuan Luu\textsuperscript{2}, Wray Buntine\textsuperscript{1}} \\
  \textsuperscript{1}VinUniversity \\
  \textsuperscript{2}Nanyang Technological University, Singapore \\
}

\begin{document}
\maketitle
\begin{abstract}
While diffusion models excel at conditionally generating high-quality images, prior works in discrete diffusion models were not evaluated on conditional long-text generation. This work addresses the limitations of prior discrete diffusion models for conditional long-text generation, particularly in the long abstractive summarization task. Despite faster decoding speeds compared to autoregressive methods, previous discrete diffusion models failed on the abstractive summarization task due to the incompatibility between the backbone architectures and the random noising process. To overcome these challenges, we introduce a novel semantic-aware noising process that enables Transformer backbones to handle long sequences effectively. Additionally, we propose CrossMamba, an adaptation of the Mamba model to the encoder-decoder paradigm, which integrates seamlessly with the random absorbing noising process. Our approaches outperform existing discrete diffusion models on three benchmark summarization datasets: Gigaword, CNN/DailyMail, and Arxiv, while also achieving much faster inference speed compared to autoregressive models. 

\end{abstract}

\section{Introduction}

Diffusion models are highly effective at generating realistic, high-quality images and have garnered considerable attention for their potential in producing discrete data types like text \cite{austin2021structured, li2021discovering, lou2024discrete}, biological sequences \cite{avdeyev2023dirichlet}, and graphs \cite{sun2023difusco, vignac2022digress}. Unlike autoregressive (AR) methods, diffusion-based models are not limited to sequential data generation, which could enhance long-term planning, controllable generation, and sampling speed. 
\begin{figure}[h!]
    \centering
    \includegraphics[width=\columnwidth]{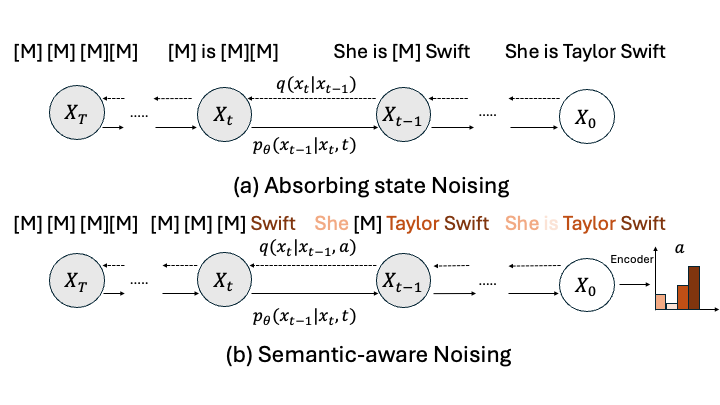}
    \caption{ In contrast to conventional discrete diffusion models, we feed the full target sequence through the encoder to obtain attention scores, reflecting the relative importance of each token to the target sentence's overall semantic meaning, and use those scores to alter the absorbing probability. The higher the attention scores, the lower the probability it is absorbed to [MASK] token, which is denoted as [M].}
    \vspace{-1em}
\end{figure}
However, discrete diffusion methods currently underperform compared to AR models \cite{austin2021structured, gulrajani2024likelihood, he2023diffusionbert, lou2024discrete}, particularly in the domain of language modeling. Recent methods aim to improve the framework by applying continuous diffusion to token embeddings \cite{gong2022diffuseq, li2022diffusion, strudel2022self, dieleman2022continuous} or logits \cite{han2022ssd, mahabadi2023tess}, necessitating complex rounding schemes to convert continuous vectors into discrete tokens. These approaches also require numerous sampling iterations, resulting in slower performance compared to autoregressive models. For example, the DiffuSeq model \cite{gong2022diffuseq} is significantly slower than a similarly scaled autoregressive baseline. Another research direction focuses on diffusion processes directly in discrete state spaces \cite{hoogeboom2022equivariant, austin2021structured,he2023diffusionbert, zheng2023reparameterized}, but this area is less explored and often produces inferior results in text generation. Consequently, despite their potential advantages in planning and controllable generation, diffusion models still face challenges in matching the efficiency and performance of autoregressive models in text generation tasks. 

Furthermore, while discrete diffusion methods theoretically could enhance the efficiency in long-sequence processing, the capability of discrete diffusion models for conditional long-text generation tasks such as abstractive summarization remains underexplored. The task of summarizing long documents presents unique complexities compared to shorter texts. Long documents often encompass multiple ideas, subtopics, and supporting details, making it challenging to identify and distill the most salient information into a coherent summary. In this work, we find out that prior works in discrete diffusion models completely fail on abstractive text summarization, as shown later in the section.~\ref{sct:exp}, due to the random absorbing noising process from D3PM \cite{austin2021structured} because the task requires a structured manner in language modeling. 

Additionally, to tackle that problem, we propose a novel forward process - A semantic-aware noising process, that utilizes the Transformer encoder-decoder architecture to force the model to generate important words first, shifting the language modeling paradigm from random to important-information-first modeling. We also introduce CrossMamba to leverage Mamba \cite{gu2023mamba} for encoder-decoder architecture,  which is well-suited for the random noising process and takes advantage of Mamba’s inherent efficiency for scaling to long sequences. By introducing the new decoding algorithm and the noising scheduler, our new framework can effectively model arbitrarily long textual sequences with linear processing time. 

In summary, our contributions are:
\begin{itemize}
    \item We introduce the problem of prior discrete diffusion frameworks in the long sequence-to-sequence task. 
    \item We propose Semantic-Aware Noising Process, a novel noise scheduler, that supports the Transformer backbone to conditionally generate long sequences in an organized manner.
    \item We propose CrossMamba, a conditioning method that leverages Mamba to encoder-decoder architecture with outstanding speed in long contexts.
    \item We conduct extensive experiments on three common abstractive text summarization benchmarks, i.e. Gigaword, CNN/DailyMail, and Arxiv, and achieve state-of-the-art results compared to other discrete diffusion models. Furthermore, our framework outperforms autoregressive and continuous diffusion models in terms of decoding time.
\end{itemize}
\section{Related Works}

\subsection{Discrete Diffusion Models}
The application of diffusion modeling to discrete data can be categorized into two main groups. The first group consists of methods that embed discrete structures into a continuous space and then apply Gaussian diffusion \cite{chen2022analog, dieleman2022continuous, gulrajani2024likelihood, han2022ssd, li2022diffusion, strudel2022self, lovelace2024latent}. 

Methods that define a diffusion process directly on discrete structures have greater potential for substantial improvements in speed.
The D3PM framework \cite{austin2021structured} introduces a Markov forward process by the multiplication of transition matrices over discrete time steps. Extending this framework to continuous time, as done in Eq.~\ref{fwd}, utilizes continuous time Markov chain (CTMC) theory \cite{campbell2022continuous}. The CTMC framework further generalizes the score-matching perspective on diffusion modeling \cite{song2019generative} to discrete data \cite{lou2024discrete, sun2022score}. Notably, SEDD \cite{lou2024discrete} integrates score-based approaches with ELBO maximization, allowing for effective likelihood-based training of score-based models.

\subsection{Abstractive Text Summarization}
Abstractive summarization involves compressing a longer input text into a shorter output summary that retains the essential information and main ideas using new phrases and sentences rather than simply extracting key phrases or sentences from the original content. Transformer-based models have dominated this field due to the ability to capture long-range dependencies and contextual relationships within the text, thanks to self-attention mechanism \cite{liu2019text, lewis2019bart, zhang2020pegasus}. However, these models fail on long abstractive summarization benchmarks due to quadratic complexity of self-attention block, which limits the number of tokens these models can handle \cite{keles2022computational}. Consequently, recent works have attempted to address this issue by incorporating new attention mechanisms \cite{guo2022longt5, zaheer2021big}. Our work tackles this problem by leveraging the linear time complexity of the Mamba model while also maintaining comparable performance with Transformer-based models on summarization benchmarks.
\begin{figure*}[!htb]
    \centering
    \includegraphics[width=\textwidth]{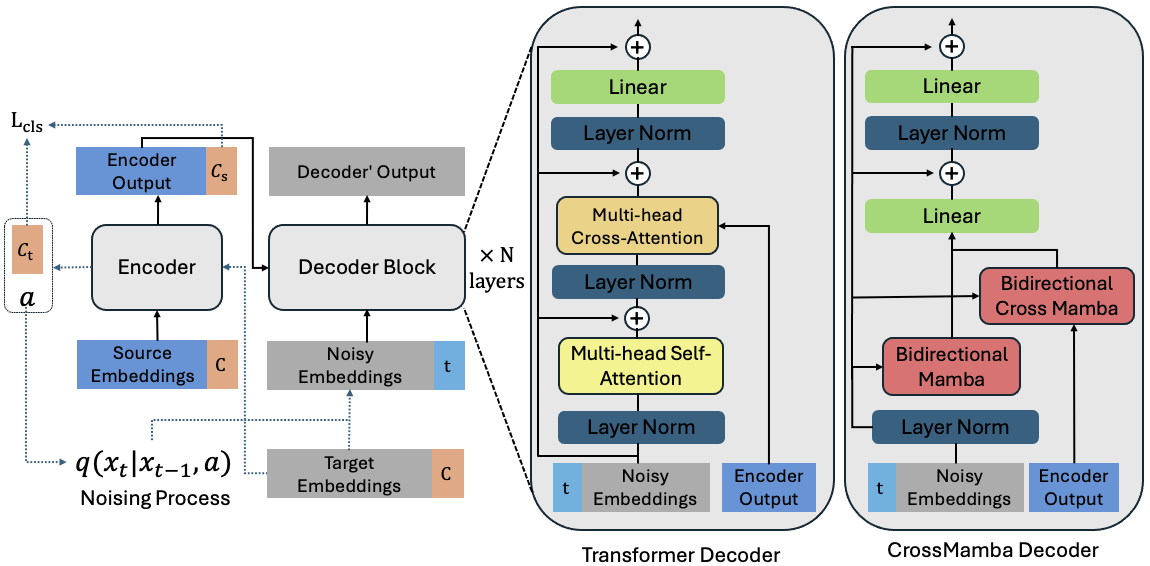}
    \caption{The model consists of an encoder and a decoder. The encoder processes the input sequence ($source$), while the decoder handles the noisy target sequence. Time step information is incorporated by adding time step embeddings $t$. The semantic-aware pipeline is illustrated by the blue dashes. A [CLS] token $C$ is appended to both the source and target sequences and then passed through the encoder. The similarity loss $L_{cls}$ is computed using the two corresponding [CLS] tokens $C_s$ and $C_t$ (detach). Additionally, the attention scores $a$ from the target sequence are used in the noising process. The decoder can be standard transformer blocks that incorporate conditioning via cross-attention or CrossMamba blocks integrating conditioning with bidirectional CrossMamba.} 
    \label{overview}
    \vspace{-1em}
\end{figure*}
\section{Methodology}
\subsection{State-Space Models}
A state-space model represents a system's dynamics using a set of input, output, and state variables defined through linear differential or difference equations involving system matrices \cite{brogan1974modern, gu2022efficiently, fu2023hungry}. The model computes the output by applying the state and input variables to the output equation involving the system matrices. Mamba \cite{gu2023mamba}, which belongs to the family of state-space models, has demonstrated significant capability in handling long sequences across a wide range of application domains. For instance, VisionMamba \cite{zhu2024vision} effectively leverages the Mamba kernel to encode images, achieving robust performance in image classification tasks. In the video domain, recent works \cite{chen2024video, liu2024vmamba} demonstrate Mamba's proficiency in managing image classification and complex spatiotemporal dynamics, offering both superior performance and favorable efficiency-performance trade-offs. In summarization task, we make the first attempt to integrate Mamba model to solve this complex language understanding task, competing with Transformer-based models. 

\vspace{-0.5em}
\subsection{Diffusion Models}

Diffusion models are trained to progressively reverse a forward corruption process \(q\) that adds noise to clean data \(\mathbf{x}\) drawn from the distribution \(q(\mathbf{x})\), generating latent variables \(\mathbf{z}_t\) for \(t \in [0, 1]\) that represent increasingly noisy versions of \(\mathbf{x}\) \cite{ho2020denoising, sahoo2023diffusion, sohl2015deep, song2020score}. The standard forward process for continuous \(\mathbf{x}\) is defined as:
\noindent
\begin{equation}
\mathbf{z}_t = \sqrt{\alpha_t} \mathbf{x} + \sqrt{1-\alpha_t} \boldsymbol{\epsilon}
\label{fwd}
\end{equation}
\noindent
where \(\boldsymbol{\epsilon} \sim \mathcal{N}(0, \mathbf{I})\) and \(\alpha_t\) is a noise schedule that decreases monotonically with \(t\). The reverse diffusion model \(p_\theta\), parameterized over \(\mathbf{x}\) and \(\mathbf{z}_t\), is trained to maximize a variational lower bound on the log-likelihood (ELBO). With \(T\) discretization steps, defining \(s(i) = \frac{(i-1)}{T}\) and \(t(i) = \frac{i}{T}\), and using \(D_{KL}[\cdot]\) to represent the Kullback-Leibler divergence, the Negative ELBO (NELBO) is given by \cite{sohl2015deep}:
\begin{align*}
L_{vb}&=\mathbb{E}_q \left[ -\log p_\theta(\mathbf{x}|\mathbf{z}_{t(0)}) \right]\\
&+ \sum_{i=1}^T D_{KL} \left[ q(\mathbf{z}_{s(i)}|\mathbf{z}_{t(i)},\mathbf{x}) \parallel p_\theta(\mathbf{z}_{s(i)}|\mathbf{z}_{t(i)}) \right] \\
&+ D_{KL} \left[ q(\mathbf{z}_{t(T)}) \parallel p_\theta(\mathbf{z}_{t(T)}) \right]
\end{align*}
\noindent
For simplicity, we omit \(i\) from \(t(i)\) and \(s(i)\) in the following discussions; generally, \(s\) will denote the time step prior to \(t\).

\subsection{Proposed Method}
RDMs \cite{zheng2023reparameterized} demonstrate that the multinominal diffusion model \cite{hoogeboom2021argmax} does not decode iteratively for further refinement, making it infeasible to generate sequences in a structured strategy. Therefore, in this study, we focus on the absorbing discrete diffusion \cite{austin2021structured}. To address the aforementioned issues of diffusion discrete Language Model for long text summarization, we (i) propose a novel forward process, the Semantic-aware Noising Process introduced in the section.~\ref{ssct:sanp}, that helps the Transformer encoder-decoder architecture overcome the limitation of conditionally generating long sequences, and (ii) develop a new backbone architecture based on Mamba, Cross-Mamba introduced in the section.~\ref{ssct:cm}, which is well-suited for the random noising process and takes advantage of Mamba's inherent efficiency for scaling to long sequences.  

Our model is broadly explained in Figure~\ref{overview}. We follow the design from SeqDiffuSeq \cite{yuan2022seqdiffuseq} promoting the encoder-decoder architecture to model the input and output text sequences. In detail, we use the encoder to process the input sequences $source$ and the decoder to model the noisy $target$ sequence. We inject time step information by adding time step embedding $t$. Using the encoder-decoder architecture offers computational convenience during generation because the input sequences ${source}$ only require one forward computation through the encoder network during the entire reverse process. Given that the reverse process requires thousands of iterations to produce high-quality output sequences, the computational resource savings can be substantial.
\subsection{Semantic Aware Noising Process}
\label{ssct:sanp}
The D3PM framework \cite{austin2021structured} introduces a Markov forward process \( q(z_t|z_{t-1}) = \text{Cat}(z_t; Q_t z_{t-1}) \) which is defined by the multiplication of matrices \( Q_t \) over \( T \) discrete time steps. This process results in the following marginal distributions:
\begin{equation*}
    q(z_t|x) = \text{Cat}(z_t; Q_t Q_{t-1} \cdots Q_1 x)  
\end{equation*} 
These marginals represent the discrete-state form of equation \ref{fwd}. Specifically, each token in the sequence either remains unchanged or transitions to [MASK] with a certain probability $\beta$. The transition matrix at time step $t$ is defined as:
\begin{equation}
\begin{aligned}
\relax [Q_t]_{ij} = 
\begin{cases} 
1 & \text{if } i = j = [M], \\
1 - \beta_{t} & \text{if } i = j \neq [M], \\
\beta_{t} & \text{if } j = [M], i \neq [M]
\end{cases}\\
\end{aligned}
\label{absorbing}
\end{equation}

As the target sequence grows longer, the random noising process makes the conditional probability of generating tokens unpredictable. In DiffusionBERT \cite{he2023diffusionbert}, the spindle noise schedule is introduced to estimate the probability that the $i$-th token remains unchanged at step $t$. This probability, denoted as $\overline{\alpha}_t^i$, is computed using the following equation $\overline{\alpha}_t^i = 1 - \frac{t}{T} - S(t) \cdot \widetilde{H}(x_o^i)$ where $\widetilde{H}$ represents the entropy, which measures the information content of a random variable, $x_i$ denotes the $i$-th token in the sequence, and $n$ denotes the length of the sequence. However, this approach requires extracting the frequencies of words in the text corpus and does not have versatility across different tasks. 

Built on top of the encoder-decoder, we feed-forward the full target sequence through the encoder yields attention scores, with the $[CLS]$ token's attention scores $[a_1, a_2, \ldots, a_n]$ indicating the relative importance of each input token to the sentence's overall semantic meaning. We reformulate the forward process equation to account for these attention scores:
\begin{equation}
\begin{aligned}
\relax [Q_t]_{ij} = 
\begin{cases} 
1 & \text{if } i = j = [M], \\
1 - P_{t} & \text{if } i = j \neq [M], \\
P_{t} & \text{if } j = [M], i \neq [M]
\end{cases}\\
\text{ with } P_{t} = \frac{t}{T} - \left(1-\frac{t}{T}\right)*a_i
\end{aligned}
\end{equation}
\noindent
with $\beta_t$ defined in Eq.\ref{absorbing}. This adjustment reflects the varying importance of different tokens at different timesteps.

Moreover, considering the semantic alignment between the input and target sequences, instead of resorting to an external pre-trained model for attention scores, both sequences are passed through the encoder. The model then calculates the cosine similarity loss between the $[CLS]$ tokens from both the source and target as:
\begin{equation}
    L_{cls} = 1 - cos(C_{s}, C_{t})
\end{equation}
\noindent
fostering end-to-end training, specifically training the encoder. This process enhances the model's semantic coherence between input and generated summaries, assuming that the two should bear a high degree of similarity. Specifically, to avoid trivial sentence embeddings, we detach $C_{t}$ from optimization. We also add the cross-entropy loss for good predictions of the data $x_0$ from $x_t$ at each time step. Thus, the total training loss is defined as:
\begin{equation}
    L_{vb} + L_{cls} + E_{q(x_0)}E_{q(x_t|x_0)}[- log \hspace{0.2em} p_{\theta}(x_0|x_t)]
    \label{loss}
\end{equation}
\noindent
\subsection{Cross-Mamba}
\label{ssct:cm}
\begin{table*}[!htb]
    \centering
    \setlength{\tabcolsep}{5pt}
    \fontsize{9}{11}
    \begin{tabular}{c|c c c c|c c c c|c c c c}
        \hline
         & \multicolumn{4}{|c}{Gigaword} & \multicolumn{4}{|c|}{CNN/DailyMail} & \multicolumn{4}{|c}{Arxiv}  \\ 
         Models & R1$\uparrow$ & R2$\uparrow$ & R-L$\uparrow$ & B $\uparrow$ & R1$\uparrow$ & R2$\uparrow$ & R-L$\uparrow$ & B $\uparrow$ & R1$\uparrow$ & R2$\uparrow$ & R-L$\uparrow$ & B $\uparrow$ \\
         \hline
         \multicolumn{13}{c}{Discrete Diffusion Models}\\
         \hdashline
         D3PM & 31.5 & 11.9 & 29.7 & 0.59 & 0.0 & 0.0 & 0.0 & 0.3 & 0.0 & 0.0 & 0.0 & 0.29\\
         DiffusionBERT & 29.3 & 9.7 & 26.1 & 0.51 & 0.0 & 0.0 & 0.0 & 0.29 & 0.0 & 0.0 & 0.0 & 0.29\\
         RDMs & 33.6 & 12.7 & 30.5 & 0.59 & 0.0 & 0.0 & 0.0 & 0.3 & 0.0 & 0.0 & 0.0 & 0.3\\
         \hline
         Semantic-aware & \textbf{37.2} & \textbf{13.2} & \textbf{35.4} & \textbf{0.65} & \textbf{32.8} & \textbf{9.5} & \textbf{29.6} & \textbf{0.56} & 0.0 & 0.0 & 0.0 & 0.3 \\
         Cross-Mamba & 35.5 & 10.6 & 33.7 & 0.63 & 23.8 & 5.3 & 21.1 & 0.51 & \textbf{21.4} & \textbf{4.3} & \textbf{20.4} & \textbf{0.46} \\ 
         \hline
         \multicolumn{13}{c}{\textcolor{gray}{Autoregressive Models}} \\
         \hdashline
         BART & 38.6 & 19.5 & 35.7 & 0.75 & 42.9 & 20.1 & 40.1 & 0.65 & 41.70 & 15.13 & 37.77 & - \\
         \hline
         \multicolumn{13}{c}{\textcolor{gray}{Continuous Diffusion Models}}\\
         \hdashline
         Tess & - & - & - & - & 41.8 & 18.3 & 35.5 & - & - & - & - & -\\
         \hline
    \end{tabular}
    \caption{Comparative analysis of various diffusion text generation models on the abstractive summarization task across Gigaword, CNN/DailyMail, and Arxiv datasets. R1, R2, RL, and B denote ROUGE-1, ROUGE-2, ROUGE-L, and bertscore, respectively. '-' indicates results are not reported in other works.}
    \label{evaluation}
    \vspace{-1em}
\end{table*}
State Space Models (SSMs) are built on continuous systems that transform a 1D function or sequence, \( x(i) \in \mathbb{R}^L \) into \( y(i) \in \mathbb{R}^L \) through an internal state \( h(i) \in \mathbb{R}^N \). Mathematically, SSMs utilize the following ordinary differential equation (ODE) to represent the input data:
\begin{align*}
    h'(i) &= Ah(i) + Bx(i)\\
    y(i) &= Ch(i)
\end{align*}
where \( A \in \mathbb{R}^{N \times N} \) is the system's evolution matrix, and \( B \in \mathbb{R}^{N \times 1}, C \in \mathbb{R}^{N \times 1} \) are the projection matrices. This continuous ODE is typically discretized in modern SSMs. Mamba \cite{gu2023mamba} represents a discrete variant of the continuous system, incorporating a timescale parameter \( \Delta \) to convert the continuous parameters \( A, B \) into their discrete forms \( \tilde{A}, \tilde{B} \). This conversion is generally done using the zero-order hold (ZOH) method, described by:
\begin{align*}
    \tilde{A} &= \exp(\Delta A) \\
    \tilde{B} &= (\Delta A)^{-1} (\exp(\Delta A) - I) \cdot \Delta B \\
    h_i &= \tilde{A} h_{i-1} + \tilde{B} x_i \\
    y_i &= Ch_i
\end{align*}

Mamba features a Selective Scan Mechanism (S6) as its primary SSM operator. The parameters \( B \in \mathbb{R}^{B \times L \times N}, C \in \mathbb{R}^{B \times L \times N}, \Delta \in \mathbb{R}^{B \times L \times D} \), are directly derived from the input data \( x \in \mathbb{R}^{B \times L \times D} \) as:
\begin{equation*}
    B, C, \Delta = s_B(x), s_C(x), s_\Delta(x)
\end{equation*}
with \( s_B(x) = \text{Linear}_N(x) \), \( s_C(x) = \text{Linear}_N(x) \), \( s_\Delta(x) = \text{Broadcast}_D(\text{Linear}_1(x)) \), and \( \tau_\Delta = \text{softplus} \), where \(\text{Linear}_d\) is a parameterized projection to dimension \(d\). The choice of \( s_\Delta \) and \( \tau_\Delta \) is motivated by their connection to RNN gating mechanisms.

Initially, we adopted a classic sequence-to-sequence RNN model, as outlined by \cite{sutskever2014sequence}, to create an encoder-decoder framework using Mamba. However, managing hidden states while maintaining rapid parallel computation proved challenging as shown in subsection~\ref{crossmamba_ablation}. 
We observed that both the self-attention \cite{vaswani2017attention} and Mamba \cite{gu2023mamba} mechanisms are input-dependent, as they generate $Key, Query, Value$ matrices and $B, C$ matrices through a linear layer, respectively. This insight led us to develop a new method called CrossMamba, which effectively addresses the information bottleneck and tailors the Mamba architecture for use in encoder-decoder models. The equations for the CrossMamba layer are expressed in equation \ref{cross}.
\begin{align}
\begin{split}
    & B_c, C_c, \Delta_c = s_B'(e_t), s_C'(e_t), s_\Delta'(e_t)\\
    &\tilde{A_c} = \exp(\Delta_c A) \\
    &\tilde{B_c} = (\Delta_c A)^{-1} (\exp(\Delta_c A) - I) \cdot \Delta_c B_c \\
    &h_i^c = \tilde{A_c} h_{i-1} + \tilde{B_c} x_i \\
    &y_i^c = C_c h_i
    \label{cross}
\end{split}
\end{align}
\noindent
with $e$ as the encoder's output. Finally, we concatenate \([y_i, y_i^c] \in \mathbb{R}^{2 \times L} \) and linear mapping the concatenation back to \(\mathbb{R}^{L}\), similar to conventional bidirectional RNN.

CMLM \cite{ghazvininejad2019mask} deploy a linear layer as a length predictor to predict the length of the target $L$ to avoid generating [PAD] tokens, and we utilize this predictor to adapt the cross-attention mechanism to create cross-Mamba. In detail, we first use Conv1d layers to compress the encoder's output according to the ratio of max source length and max target length. Let $N$ be the length of the encoder's output after compression, if $N<L$, we pad the sequence to the same length $L$; otherwise, we take the last $L$ tokens from the encoder's output to create the matrices $B_c$ and $C_c$. The two matrices $B_c$ and $C_c$ are used to compute the target sequence in equation \ref{cross}.

\section{Experiments}
\label{sct:exp}
We evaluate our model on various sequence-to-sequence benchmarks and focus on text summarization datasets, including Gigaword \cite{rush-etal-2015-neural}, CNN/DailyMail (CNNDM) \cite{nallapati2016abstractive}, and Arxiv \cite{cohan-etal-2018-discourse}. We also compare the decoding speed of our models with autoregressive models. 

\subsection{Results}
Our quantitative results are presented in Table~\ref{evaluation}, showcasing ROUGE-1 (unigram), ROUGE-2 (bigram), ROUGE-L (longest common subsequence) scores, and bertscore following prior text summarization work \cite{lewis2019bart}. A comprehensive table of evaluation results can be found in appendix~\ref{full_eval}. Generally, all previous discrete diffusion models are unable to generate sequences conditionally for the CNN/DailyMail dataset. In contrast, our proposed methods significantly outperform them, achieving improvements of up to 32 and 30 points in ROUGE-1 and ROUGE-L scores, respectively. Although semantic-aware noising continues to struggle with the ArXiv dataset, our Cross-Mamba method maintains consistent performance, achieving respectable scores of 21.4 in ROUGE-1 and 20.4 in ROUGE-L. In a simpler text summarization dataset like Gigaword, the semantic-aware method still outperforms across all four metrics, implying that our methods not only possess the capability to generate longer sequences but also produce higher-quality outputs.

\subsubsection{Human Evaluations}
We conduct human evaluations to examine the outputs generated by the model. Specifically, we evaluate the outputs from DiffusionBERT, RDMs, our framework, and the gold standard summaries across four categories: Relevance, Consistency, Fluency, and Coherence. Each category is assessed using a five-point Likert scale, where scores range from 1 to 5 (worst to best). The Gigaword dataset is used for the experiment. We randomly selected 50 output samples and asked 5 professional English speakers to evaluate them. The mean score for each category of each model is reported in Table~\ref{human_eval}. 
Additionally, \(p\) denotes the Spearman correlation between annotators, reflecting the agreement among all annotators on the final scores. As shown in the table, the results indicate good inter-annotator agreement, with an average correlation of 0.79 across all categories. Our framework outperforms the other models on every evaluated criterion. It achieves Relevance and Consistency scores of 4.15 and 4.31, respectively, significantly surpassing the next-best Semantic-Aware model, which scored 3.41 and 3.63. Furthermore, our model scores 3.9 and 4.44 on the Fluency and Coherence tests, demonstrating its strong capability in handling the summarization task with performance comparable to the reference.
\begin{table*}[!htb]
    \centering
    \fontsize{9}{11}
    \begin{tabular}{c|c c|c c|c c|c c}
         Models & \multicolumn{2}{|c}{Relevance} & \multicolumn{2}{|c}{Consistency} & \multicolumn{2}{|c}{Fluency} & \multicolumn{2}{|c}{Coherence} \\
         & Mean & p & Mean & p & Mean & p & Mean & p \\
         \hline
         DiffusionBERT & 2.61 & 0.60 & 2.91 & 0.83 & 3.09 & 0.88 & 3.06 & 0.79\\
         RDMs & 2.82 & 0.79 & 3.2 & 0.63 & 3.11 & 0.58 & 3.15 & 0.75\\
         Semantic-Aware & 3.41 & 0.84 & 3.63 & 0.86 & 3.46 & 0.85 & 3.61 & 0.60\\
         \hline
         Reference & 4.15 & 0.59 & 4.31 & 0.87 & 3.9 & 0.71 & 4.44 & 0.69\\
         \hline
    \end{tabular}
    \caption{Comparison of models based on Relevance, Consistency, Fluency, and Coherence, as evaluated by humans on the Gigaword dataset. Reference refers to the human annotations, and \(p\) denotes the Spearman correlation.}
    \label{human_eval}
    \vspace{-1em}
\end{table*}

\subsubsection{{Decoding Speed}}
This section presents a performance-runtime comparison of various text generation models. Specifically, the BART decoder is causal, meaning that generation depends on the length of the target sequences rather than a constant number of steps. Continuous diffusion models typically require training with up to \(T = 2000\) diffusion steps, resulting in a need for a minimum of \(T > 50\) \cite{wu2023ar} sampling steps to achieve good performance on the CNN/DM dataset.

By incorporating features from other discrete diffusion models and leveraging the efficiency of Mamba, our model achieves exceptional decoding speed on the CNN/DailyMail dataset, significantly outperforming autoregressive models. As shown in Table \ref{speed}, with just 10 inference steps, our model with CrossMamba runs up to 4 times faster than both BART and TESS, while the Semantic-aware method is 2 times faster. Despite having 50 diffusion timesteps for training, both CrossMamba and Semantic-aware can still deliver impressive results with only 2 inference steps, achieving speeds up to 15 times and 8 times faster than BART, respectively. In contrast, TESS experiences a marginal performance decline as the number of steps decreases from 100 to 10, and Genie's R-L performance drastically drops when the inference steps are reduced from 1000 to 100.
\begin{table}[!htb]
    \centering
    \begin{tabular}{c| c c c}
        & Step & Speed  & R-L \\ 
        \hline
         BART & n/a &  1.00 & 40.1\\
         \hdashline
         TESS & 100 & 0.92 & 35.6 \\
         TESS & 1000 & 0.11 & 39.7 \\
         \hdashline
         Semantic-aware & 2 & 7.92 & 27.5 \\
         Semantic-aware & 10 & 2.10 & 29.6\\
         CrossMamba & 2 &\textbf{15.20} & 19.7\\
         CrossMamba & 10 & 4.10 & 21.1\\ 
         \hline 
    \end{tabular}
    \caption{Decoding speed relative to BART (expressed as a multiplier) for two backbone architectures with different numbers of diffusion timesteps, reported on the CNN/DailyMail dataset.}
    \label{speed}
    \vspace{-1em}
\end{table}

\subsection{Analysis}

In this section, we study how the semantic-aware noising process influences both the decoding stage and the training stage.
\subsubsection{Effect of Semantic-aware Noising}

In the summarization task, the target should encapsulate the core meaning according to the source sequence. Therefore, minimizing the similarity loss between the source and target sequence will ensure the consistency between the source input and the generated sequence of the model. This will signal the model to produce more concise sequences, including accurately identifying and generating correct entities (such as persons, objects, etc.). As demonstrated in Table \ref{qualitative}, the model consistently generates important words first, specifically named entities, thereby highlighting the efficacy of the semantic-aware noising process.

To shed light on the stagnant performance of the semantic-aware method on the ArXiv dataset, we compare the entropy scores of the noising distribution \(Q_t\). The more uniform the distribution, the higher the entropy score, with the maximum entropy value being \(\log_2 N\), where \(N\) is the sequence length.
\begin{table}[!htb]
    \centering
    \begin{tabular}{c|c c}
         Dataset & E & max E\\
         \hline
         CNN/DM & 3.56 & 8 \\
         Arxiv & 8.71 & 10 \\
         \hline
    \end{tabular}
    \caption{Entropy scores, denoted as \(E\), computed from \(Q_t\), express the uniformity of the distribution, and \(\text{max } E\) represents the maximum value when \(Q_t\) is perfectly uniform.}
    \label{uniform}
\end{table}
\noindent
Table~\ref{uniform} illustrates that the uniformity of \(Q_t\) in the ArXiv dataset is significant, nearing the maximum, which hinders the construction of an organized decoding stage. In contrast, the entropy score of \(Q_t\) in the CNN/DM dataset is slightly lower, indicating less uniformity. This difference arises because the target sequences still contain many tokens with identical attention scores, which do not contribute much to the overall semantic meaning of the sequences.


\begin{table*}[!htb]
    \centering
    \fontsize{10}{11}\selectfont
    \begin{tabular}{c | c}
        \hline
        \multirow{2}{*}{t = 2} &  [M] [M] May [M] [M] [M] [M] [M] [M] [M] night. [M] Pacquiao will [M] [M] [M] [M] \\
        & [M] [M] [M] [M] [M]. [M] [M] [M] [M] [M] [M] [M] fight on [M] [M] [M] [M] [M] [M]\\ [0.3em]
        \hdashline
        \multirow{2}{*}{t = 5} & Floyd Mayweather will [M] at the [M] in [M]. He is a [M] [M] [M] [M]. the [M] [M] [M] [M]\\ 
        &[M] [M] [M] Pacquiao [M] [M] May [M] [M] [M].  M] [M] here for the [M] [M] the news [M] [M]\\ [0.3em]
        \hdashline
        \multirow{2}{*}{t = 10} & Floyd Mayweather will start at the gym in May. He is a four-time trainer. the Filipino is \\
        & currently for the night. Manny Pacquiao on May 11. Click here for the latest of the news.\\
        \hline
    \end{tabular}
    \caption{Generation of the Transformer encoder-decoder model trained with the Semantic-aware Noising over time. The input is from the CNN/DailyMail dataset, with [M] representing the [MASK] token. In the examples, the model first generates important words, such as named entities (Floyd Mayweather, Manny Pacquiao).}
    \label{qualitative}
    \vspace{-1.5em}
\end{table*}
\section{Ablation Studies}
In this section, we conduct ablation studies on the effect of the similarity loss, detaching the target's $[CLS]$ token as well as the design choice of CrossMamba. 
\subsection{Cross-Mamba Layer}
\label{crossmamba_ablation}
To understand more about the design of CrossMamba, we compared it with other prominent techniques that utilize RNN-based models, including seq2seq and Diffuseq. We chose the QQP dataset for this experiment because the paraphrasing task it presents is simpler compared to tasks like summarization. Table \ref{crossmamba} demonstrates that our method excels at connecting the source and target sequences, and almost matches the attention mechanism whereas seq2seq suffers from an information bottleneck problem, and Diffuseq requires the model to reconstruct the input.
\begin{table}[!htb]
    \centering
    \begin{tabular}{c|c c c}
         & BLEU & R-L & bertscore  \\
         \hline
         CLS seq2seq & 8.3 & 28 & 0.62 \\ 
         Diffuseq & 16.5 & 48 & 0.75 \\
         \hline 
         CrossMamba & \textbf{21.2} &\textbf{56.4} & \textbf{0.81} \\
         \hline
         BART & - & - & 0.67 \\
         \hline
    \end{tabular}
    \caption{Different approaches adapting Mamba to discrete diffusion models on simple QQP paraphrasing dataset, showing that CrossMamba outperforms other Seq2Seq RNN techniques.}
    \label{crossmamba}
    \vspace{-1em}
\end{table}

Intuitively, the attention mechanism computes a categorical distribution from \(K, Q, V\) across the sequence, whereas Mamba's \(B\) and \(C\) matrices are derived from the corresponding input tokens and encapsulate the sequence information into hidden states. Therefore, we hypothesize that Mamba's kernels are more independent than the attention kernel, enabling it to perform better during random noise processing. 
\begin{table}[!htb]
    \centering
    \begin{tabular}{c|c c c}
        & R-1 & R-2 & R-L \\
        \hline
         Transformer-CrossMamba & 15.8 & 3.1 & 14.7 \\
         Mamba-CrossAttention & 15.1 & 2.9 & 14.0 \\
         \hline
         Mamba-CrossMamba & \textbf{23.8} & \textbf{5.3} & \textbf{21.1} \\
         \hline 
    \end{tabular}
    \caption{Quantitative results on different combinations of Mamba and Transformers on CNN/DailyMail dataset. The left model is the Encoder and the right model is the Decoder.}
    \label{design}
\end{table}
\noindent
To test this hypothesis, we trained two different combinations of Mamba and attention mechanisms. First, we replaced CrossMamba in the Mamba decoder with cross-attention. Second, we tested a Transformer encoder with a CrossMamba decoder. Our results, shown in Table \ref{design}, demonstrate that both configurations underperform in handling noise compared to the Mamba encoder - CrossMamba decoder setup. This suggests that the attention mechanism is incompatible with the random noise processing scenario.
\subsection{Effect of Similarity Loss}

\textbf{Without Similarity Loss:} Without the similarity loss, there is no guarantee that the attention scores are consistent with the semantic meaning of the target and the noising process remains random, failing to dismantle the sequence in a structured manner. As shown in \ref{cls_loss}, removing similarity loss causes R-1 score drops by 6.6 points, R-2 score drops by 3.8 points, and R-L score drops by 5.8 points 
\begin{table}[!htb]
    \centering
    \begin{tabular}{c|c c c}
         & R-1 & R-2 & R-L \\
         \hline
        Removing & 26.2 & 5.7 & 23.8 \\
        Non-detach & 26.9 & 5.5 & 24.6 \\
        \hline 
        Semantic-aware & \textbf{32.8} & \textbf{9.5} & \textbf{29.6} \\
        \hline
    \end{tabular}
    \caption{Result of the semantic-aware noising on CNNDM dataset without the similarity loss and non-detach target sequence scenarios}
    \vspace{-1em}
    \label{cls_loss}
\end{table}

\noindent\textbf{Not Detach target sequence: }Compute the gradient on both the source's $[CLS]$ and the target's $[CLS]$ shift the sequence-to-sequence task to classification, and the model can reach a trivial solution for sentence embedding, and a tremendous decrease in all metrics  as illustrated in Table \ref{cls_loss}. In detail, there are marginal reductions of 5.9, 4.0, 5.0 in R-1, R-2, and R-L, respectively. These empirical evidences highlight substantial performance gains provided by semantic-aware noising.
\section{Conclusion}
We introduce the Semantic-Aware Noising Process, a noise scheduler for Transformers that enables structured conditional generation of long sequences. Additionally, CrossMamba enhances encoder-decoder architectures for handling long contexts with exceptional speed. Our approach achieves state-of-the-art results on summarization benchmarks like Gigaword, CNN/DailyMail, and Arxiv, while surpassing autoregressive and continuous diffusion models in decoding speed, advancing discrete diffusion models for long-context generation.
\newpage
\section{Limitations}

We have presented the Semantic-aware noising process and CrossMamba to tackle the main limitation of discrete diffusion models in conditional long-context sequences processing. We achieve strong empirical results relative to previous works on discrete diffusion models but still drop behind Autoregressive Models. One significant limitation is the suboptimal performance of the noising scheduler, which may be attributed to the trainability of the encoder. This issue suggests that more advanced techniques, such as distillation methods, could potentially enhance the encoder's effectiveness and overall model performance. Exploring these methods could be a promising direction for future work. Another challenge we identified is the scalability of the proposed noising scheduler. While it shows promise, it struggles with very long sequences, such as those found in the Arxiv dataset. Future research could focus on developing a more structured noising scheduler that can handle longer sequences more efficiently, such as adapting the attention weights only to the most important tokens.

\bibliography{custom}

\appendix
\label{sec:appendix}
\clearpage
\section{Evaluations}
\label{full_eval}
\begin{table*}[!htb]
    \centering
    \setlength{\tabcolsep}{5pt}
    \fontsize{9}{11}
    \begin{tabular}{c|c c c c|c c c c|c c c c}
        \hline
         & \multicolumn{4}{|c}{Gigaword} & \multicolumn{4}{|c|}{CNN/DailyMail} & \multicolumn{4}{|c}{Arxiv}  \\ 
         Models & R1$\uparrow$ & R2$\uparrow$ & R-L$\uparrow$ & B $\uparrow$ & R1$\uparrow$ & R2$\uparrow$ & R-L$\uparrow$ & B $\uparrow$ & R1$\uparrow$ & R2$\uparrow$ & R-L$\uparrow$ & B $\uparrow$ \\
         \hline
         \multicolumn{13}{c}{Discrete Diffusion Models}\\
         \hdashline
         D3PM & 31.5 & 11.9 & 29.7 & 0.59 & 0.0 & 0.0 & 0.0 & 0.3 & 0.0 & 0.0 & 0.0 & 0.29\\
         DiffusionBERT & 29.3 & 9.7 & 26.1 & 0.51 & 0.0 & 0.0 & 0.0 & 0.29 & 0.0 & 0.0 & 0.0 & 0.29\\
         RDMs & 33.6 & 12.7 & 30.5 & 0.59 & 0.0 & 0.0 & 0.0 & 0.3 & 0.0 & 0.0 & 0.0 & 0.3\\
         \hline
         Semantic-aware & \textbf{37.2} & \textbf{13.2} & \textbf{35.4} & \textbf{0.65} & \textbf{32.8} & \textbf{9.5} & \textbf{29.6} & \textbf{0.56} & 0.0 & 0.0 & 0.0 & 0.3 \\
         Cross-Mamba & 35.5 & 10.6 & 33.7 & 0.63 & 23.8 & 5.3 & 21.1 & 0.51 & \textbf{21.4} & \textbf{4.3} & \textbf{20.4} & 0.46 \\ 
         \hline
         \multicolumn{13}{c}{\textcolor{gray}{Autoregressive Models}} \\
         \hdashline
         BART & 38.6 & 19.5 & 35.7 & & 42.9 & 20.1 & 40.1 & 0.65 & 41.70 & 15.13 & 37.77 & - \\
         \hline
         \multicolumn{13}{c}{\textcolor{gray}{Continuous Diffusion Models}}\\
         \hdashline
         Tess & - & - & - & - & 41.8 & 18.3 & 35.5 & - & - & - & - & -\\
         AR-Diffusion & - & - & - & - & 40.2 & 17.1 & 37.7 & - & - & - & - & -\\
         GENIE & 45.7 & 25.8 & 42.9 & - & 45.6 & 23.2 & 43.1 & - & - & - & - & -\\
         \hline
    \end{tabular}
    \caption{Extensive analysis of various diffusion text generation models on the abstractive summarization task across Gigaword, CNN/DailyMail, and Arxiv datasets. R1, R2, RL, and B denote ROUGE-1, ROUGE-2, ROUGE-L, and bertscore, respectively. '-' indicates results are not reported in other works.}
    \label{evaluation_full}
    \vspace{-1em}
\end{table*}
We include the full benchmark in Table~\ref{evaluation_full}.

\section{Implementation Details}
We set the number diffusion timestep $T$ in training to $T=50$ and inference for evaluation to $T=10$. We construct the encoder and decoder with 8 layers for each. Our model with the Transformer backbone has about 90M parameters and the Mamba backbone has roughly 85M parameters. We train the model using the AdamW optimizer \cite{loshchilov2017decoupled} for 100,000 training steps, with a learning rate of $5 \times 10^{-5}$. During the initial 10,000 steps, we employ a linear warmup schedule starting from a learning rate of $5 \times 10^{-8}$. All experiments are conducted on 2 NVIDIA RTX 3090 GPUs and we use 1 for sampling. Our implementation is also based on $FairSeq$ toolkit \cite{ott2019fairseq} like RDMs \cite{zheng2023reparameterized}.

\section{Convergence Speed}
\begin{figure}[!htb]
    \centering
    \includegraphics[width=0.8\columnwidth]{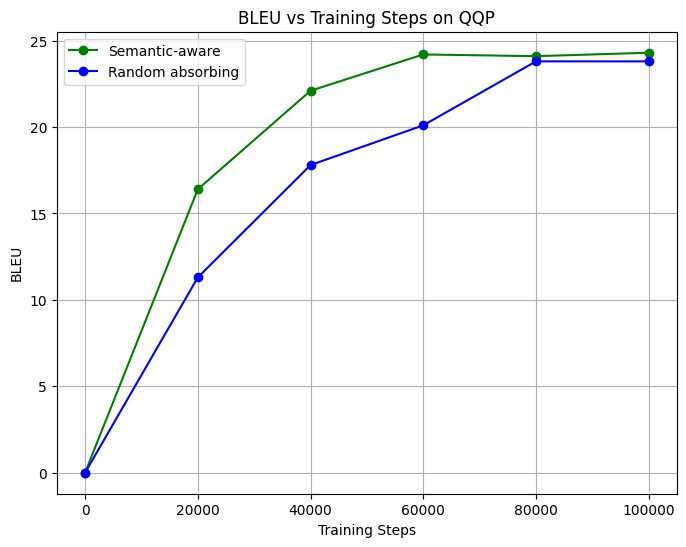}
    \caption{Curves of BLEU score vs training steps on the QQP dataset with absorbing noising and semantic-aware noising.}
    \vspace{-1em}
    \label{converge}
\end{figure}

Figure \ref{converge} demonstrates that with the implementation of semantic-aware noising, the training process converges significantly faster on the QQP dataset compared to D3PM using random absorbing. At 20,000 training steps, the semantic-aware noising scheduler demonstrates performance comparable to that of a random noising scheduler trained for 40,000 steps. Furthermore, at 40,000 training steps, it surpasses the random noising scheduler trained on 60,000 steps by a large margin in terms of BLEU score on the QQP dataset. This finding suggests that discrete diffusion models can achieve enhanced performance through the development of appropriate generation strategies.

\end{document}